\pdfoutput=1
\PassOptionsToPackage{dvipsnames}{xcolor} 
\documentclass[11pt]{article}

\usepackage[]{acl}

\ExplSyntaxOn
\NewDocumentCommand{\xhrulefill}{O{}}
 {
  \group_begin:
  \severin_xhrulefill:n { #1 }
  \group_end:
 }

\keys_define:nn { severin/xhrulefill }
 {
  height .dim_set:N    = \l_severin_xhrule_height_dim,
  thickness .dim_set:N = \l_severin_xhrule_thickness_dim,
  fill .skip_set:N     = \l_severin_xhrule_fill_skip,
  height .initial:n    = 0pt,
  thickness .initial:n = 0.4pt,
  fill .initial:n      = 0pt plus 1fill,
 }

\cs_new_protected:Nn \severin_xhrulefill:n
 {
  \keys_set:nn { severin/xhrulefill } { #1 }
  \leavevmode
  \leaders\hrule 
    height \dim_eval:n { \l_severin_xhrule_thickness_dim + \l_severin_xhrule_height_dim }
    depth  \dim_eval:n { -\l_severin_xhrule_height_dim }
  \skip_horizontal:N \l_severin_xhrule_fill_skip
  \kern 0pt
}
\ExplSyntaxOff

\usepackage{times}
\usepackage{latexsym}
\usepackage[utf8]{inputenc}
\usepackage[T1]{fontenc}
\usepackage{CJKutf8}
\usepackage[english]{babel}
\usepackage{makecell}
\usepackage{graphicx}
\usepackage{hyperref}
\usepackage{multirow} 
\usepackage{float} 
\usepackage{amsmath} 
\usepackage{adjustbox} 
\usepackage{verbatim} 
\usepackage{hwemoji}
\usepackage{pifont} 
\usepackage[normalem]{ulem}
\useunder{\uline}{\ul}{}
\usepackage[dvipsnames]{xcolor} 
\usepackage{tabularx}

\usepackage[T1]{fontenc}

\usepackage[utf8]{inputenc}

\usepackage{microtype}

\title{\textsc{Multi-News\textsuperscript{+}}: Cost-efficient Dataset Cleansing \\ via LLM-based Data Annotation}

\author{Juhwan Choi\textsuperscript{1}, Jungmin Yun\textsuperscript{1}, Kyohoon Jin\textsuperscript{2} \and YoungBin Kim\textsuperscript{1,2} \\\\
  \textsuperscript{1}Department of Artificial Intelligence, Chung-Ang University \\
  \textsuperscript{2}Graduate School of Advanced Imaging Sciences, Multimedia and Film, Chung-Ang University \\
  \texttt{\{gold5230, cocoro357, fhzh123, ybkim85\}@cau.ac.kr} \\
}

\begin{document}
\maketitle

\begin{abstract}
The quality of the dataset is crucial for ensuring optimal performance and reliability of downstream task models. However, datasets often contain noisy data inadvertently included during the construction process. Numerous attempts have been made to correct this issue through human annotators. However, hiring and managing human annotators is expensive and time-consuming. As an alternative, recent studies are exploring the use of large language models (LLMs) for data annotation. 

In this study, we present a case study that extends the application of LLM-based data annotation to enhance the quality of existing datasets through a \textit{cleansing} strategy. Specifically, we leverage approaches such as chain-of-thought and majority voting to imitate human annotation and classify unrelated documents from the Multi-News dataset, which is widely used for the multi-document summarization task. Through our proposed cleansing method, we introduce an enhanced \textsc{Multi-News\textsuperscript{+}}. By employing LLMs for data cleansing, we demonstrate an efficient and effective approach to improving dataset quality without relying on expensive human annotation efforts.
\end{abstract}

\begin{figure*}[t]
    \centering
    \includegraphics[width=\textwidth]{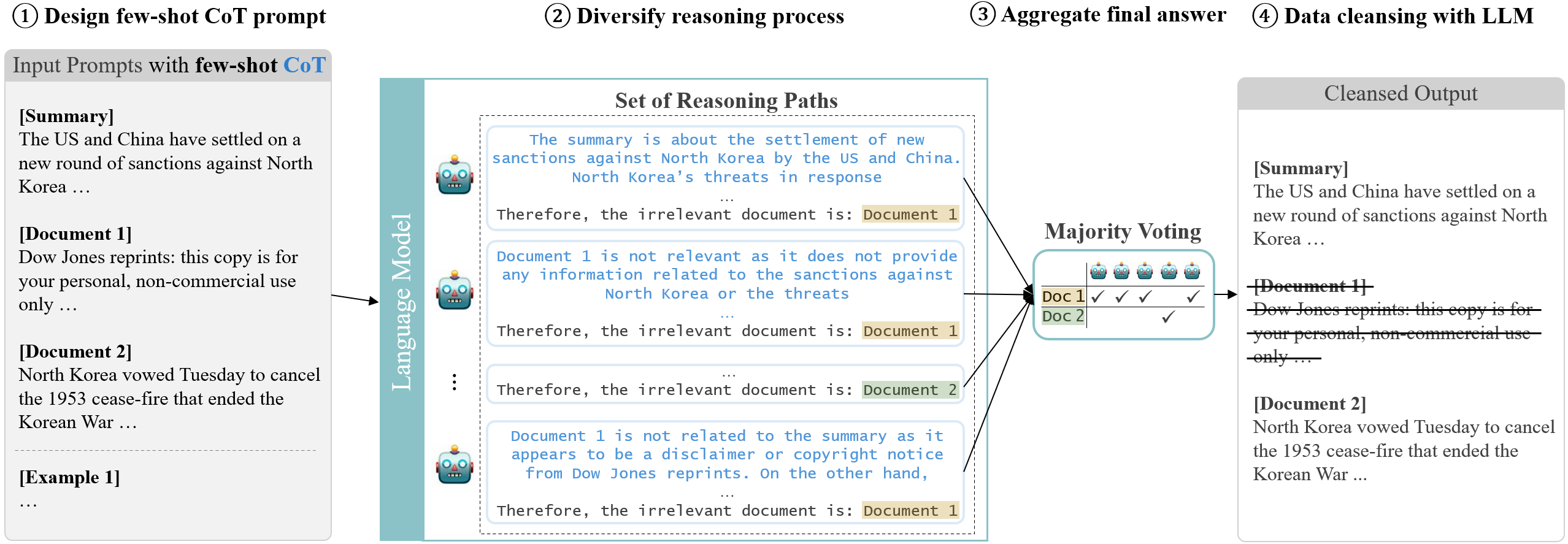}
    \caption{Overall framework for cleansing data and composing \textsc{Multi-News\textsuperscript{+}}.}
\label{fig-framework}
\end{figure*}

\section{Introduction}
 
The significance of dataset quality in deep learning applications cannot be overstated as mislabeled or noisy data can severely degrade performance \cite{song2022learning}. Datasets with incorrect labels, noise, or inconsistencies undermine the consistency and stability of model training. Cleansing these datasets contributes to enhancing model performance and generalization capabilities. Hence, ensuring the quality of the dataset by identifying and eliminating noisy data is imperative. In the realm of natural language processing, several researchers have attempted to improve the quality of noisy datasets \cite{jiang2020neural, jiang2022arxivedits}. For example, ReDocRED \cite{tan2022revisiting} addressed issues such as false negatives in DocRED \cite{yao2019docred}, a widely used dataset for relation extraction. Similarly, annotation inconsistencies were found in the MultiWOZ dataset \cite{budzianowski2018multiwoz} for dialogue state tracking \cite{qian2021annotation}, leading to efforts to rectify these issues \cite{eric2020multiwoz, zang2020multiwoz, han2021multiwoz, ye2022multiwoz}.
  
\begin{table}[t!]
\centering
\small
\begin{tabularx}{\columnwidth}{|X|}
\hline
\textbf{Source 1} \\ \hline
\textcolor{gray}{\sout{Starting in 1996, alexa internet has been donating their crawl data to the internet archive. Flowing in every day, these data are added to the wayback machine after an embargo period.}}\\

\hline

\textbf{Source 2} \\ \hline
... For the first time in decades, \textcolor{red}{researchers trying to develop a vaccine for malaria have discovered a new target} they can use to attack this deadly and common parasite...\\

\hline
\textbf{Source 3}\\ \hline
\textcolor{gray}{\sout{Focused crawls are collections of frequently-updated webcrawl data from narrow ( as opposed to broad or wide ) web crawls, often focused on a single domain or subdomain.}}\\
\hline
\hline
\textbf{Summary} \\ \hline
\textcolor{red}{Researchers think they've found a promising new potential weapon in the fight against malaria} in a fairly unlikely place: the blood of toddlers. In a paper published in science today, ...\\
\hline
\end{tabularx}
\caption{Examples of noisy documents in Multi-News dataset. Sources 1 and 3 do not contribute to the summary. We aim to identify such noisy documents without a human annotator.}
\label{tab-example}
\end{table}

Despite these efforts, relying on human annotators to enhance datasets poses challenges such as high costs and time constraints. The quality of the annotation might also be affected by potential variations, such as subjective bias and the proficiency of the annotator~\cite{rashtchian2010collecting}. Furthermore, cleansing a noisy dataset typically requires a larger budget, often involving majority voting by multiple annotators or validation by experts \cite{tan2022revisiting}. Given the significance and necessity of enhancing the quality of existing datasets, these obstacles hinder practical efforts to cleanse datasets efficiently. Therefore, it is crucial to explore cost-effective methods that can cleanse the existing dataset, minimizing human involvement.
 
In this study, we propose leveraging large language model (LLM)-based annotation for dataset cleansing. Researchers have explored cost-efficient alternatives to human annotators by employing LLMs across various tasks \cite{wang2021want, ding2023gpt, he2024annollm, bansal2023large, zhang-etal-2023-llmaaa, choi2024gpts}. However, the real-world applicability of LLM-based annotation on existing datasets is still less explored. Building on these insights, we extend the application of LLM-based annotations to denoise the existing dataset and improve its quality. Specifically, we conduct a case study to cleanse the Multi-News~\cite{fabbri2019multi}, a dataset for multi-document summarization tasks. This dataset consists of news articles crawled from the internet and is widely used in multi-document summarization research. However, as shown in Table~\ref{tab-example}, we identify several issues related to the noise in the dataset. For instance, the set of documents contained system messages from platforms such as Twitter, Wayback Machine, or Dow Jones that are unrelated to the summary and degrade the dataset quality.
  
To accomplish our purpose, we utilize LLMs to analyze the summary and associated documents, identifying and excluding any documents that are not relevant to the summary. Specifically, we employ approaches such as chain-of-thought (CoT), providing the rationale for decision-making with enhanced transparency and facilitating human investigation. We further enhance our cleansing process by incorporating self-consistency considerations, which mimic the majority voting process used by human annotators~\cite{wang2022self}. Based on our carefully designed framework, we introduce \textsc{Multi-News\textsuperscript{+}}, an enhanced version of the existing Multi-News dataset, achieved through our LLM-based cleansing strategy. To the best of our knowledge, this is the first attempt to exploit LLMs to enhance the quality of real-world datasets. Our experiments demonstrate the effectiveness of \textsc{Multi-News\textsuperscript{+}}, providing a valuable resource for future research. We make \textsc{Multi-News\textsuperscript{+}} and our source code publicly available for further study.

\section{Related Work}
Dataset quality has been an interest to researchers because of its importance in ensuring the quality of the model trained with the dataset \cite{budach2022effects}. Previous studies found that large amounts of data automatically crawled from the web may contain noisy documents, and proper filtering procedures can be an efficient solution against them \cite{xu2017zipporah, khayrallah2018impact, kryscinski2019neural, luccioni2021s, kreutzer2022quality}. Accordingly, several studies in text summarization investigated various strategies to filter out noisy data \cite{matsumaru2020improving, nan2021entity, guo2022questioning} and released new datasets with better quality \cite{grusky2018newsroom, urlana2022tesum}. However, their strategies are primarily composed of coarse rule-based methods and less interpretable model output, or costly human investigation has been applied for constructing new datasets. Furthermore, such strategies have not been applied to multi-document summarization datasets.

In the meantime, with the advancement of LLMs \cite{zhao2023survey}, researchers have explored the usage of LLMs for data annotation, a task that traditionally relied on human annotators. Initial attempts have revealed the potential capabilities of models like GPT-3 for data annotation \cite{wang2021want}. These studies indicate that GPT-3 can annotate datasets more efficiently and cost-effectively than human annotators. This results in enhanced downstream task performance, with the model trained on the GPT-3 annotated dataset outperforming the one trained on the human-annotated dataset. Subsequent studies have further demonstrated the capabilities of GPT-3, showing its ability to generate labeled data using external knowledge or instructions about desired labels and domains \cite{ding2023gpt}. Additionally, researchers have examined the usefulness of newer models like GPT-3.5 and evaluated the effectiveness of CoT in improving annotation quality \cite{he2024annollm}. LLM-based annotation has also been extended to low-resource languages where hiring human annotators is challenging \cite{choi2024gpts}.
 
In this work, we introduce a novel approach to filtering noisy documents from multi-document summarization dataset by extending cost-efficient LLM-based annotation beyond traditional data annotation tasks. By leveraging the capabilities of LLMs, our study facilitates real-world dataset cleansing, enhancing the quality of existing datasets. This attempt is noteworthy as it broadens the scope of LLM applications, offering effective solutions for improving dataset quality and streamlining its cleansing process, minimizing reliance on human annotations.

\section{\textsc{Multi-News\textsuperscript{+}}}
\label{sec:method}

\begin{figure}[t!]
    \centering
    \includegraphics[width=\columnwidth]{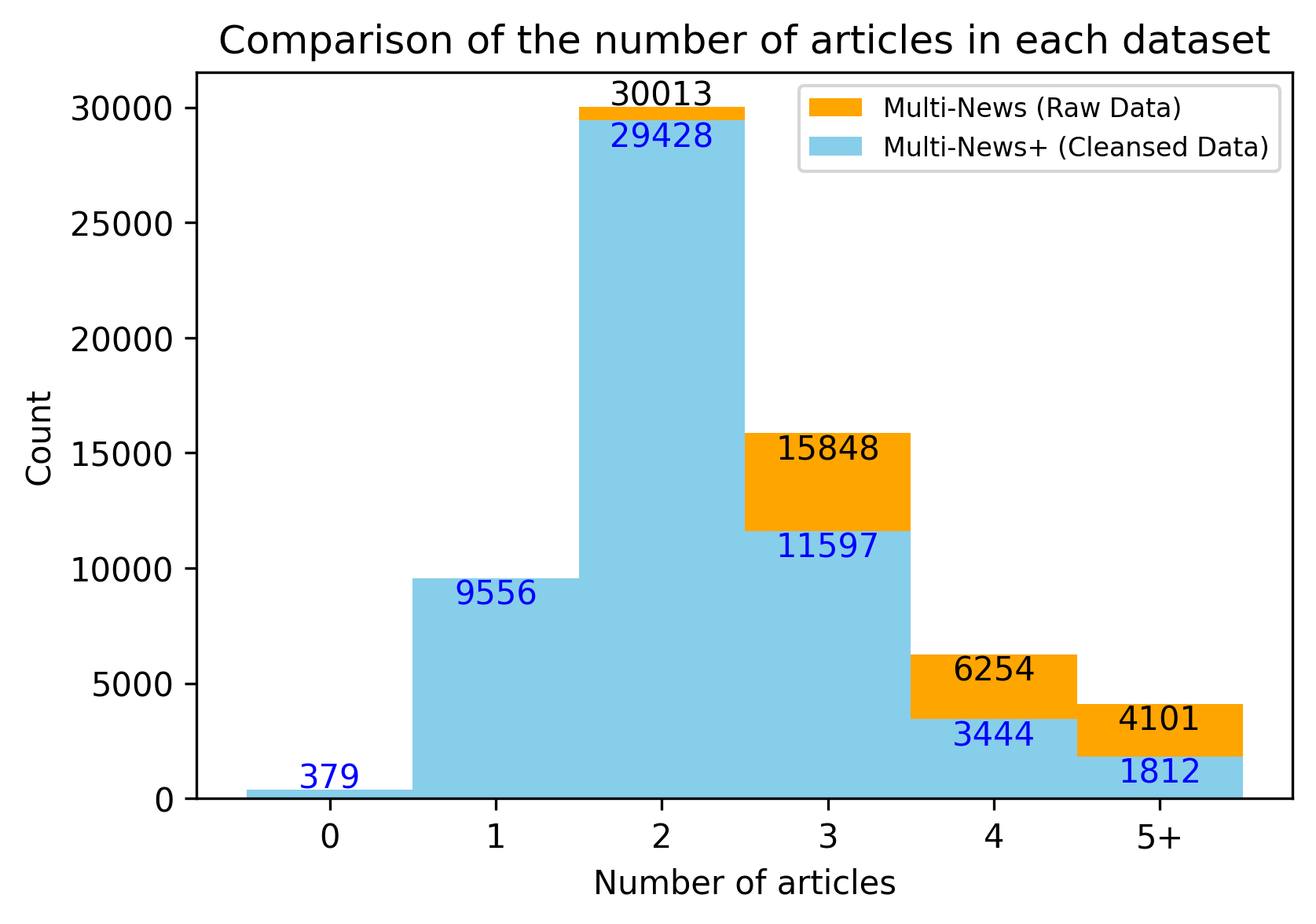}
    \caption{Histogram comparing the amount of input articles in each dataset.}
\label{fig-histogram}
\end{figure}

\begin{table*}[t!]
\centering
\resizebox{0.8\textwidth}{!}{%
\begin{tabular}{c|ccccc}
\Xhline{3\arrayrulewidth}
Model                                  & \multicolumn{5}{c}{\textit{BART-large-cnn}}                                          \\
Metric                                 & ROUGE-1        & ROUGE-2        & ROUGE-L        & BERTScore       & BARTScore       \\ \hline\hline
Multi-News                             & 48.64          & 18.86          & 24.11          & 0.6401          & -2.763          \\
\textsc{Multi-News\textsuperscript{+}} & \textbf{49.17} & \textbf{19.04} & \textbf{24.36} & \textbf{0.6418} & \textbf{-2.698} \\ 
Ablation \cite{urlana2022tesum}        & 47.48          & 18.27          & 23.81          & 0.6362          & -2.767          \\ \Xhline{2\arrayrulewidth}
Model                                  & \multicolumn{5}{c}{\textit{T5-base}}                                \\
Metric                                 & ROUGE-1        & ROUGE-2        & ROUGE-L        & BERTScore       & BARTScore       \\ \hline\hline
Multi-News                             & 40.11          & 13.90          & 21.58          & 0.6003          & -2.407          \\
\textsc{Multi-News\textsuperscript{+}} & \textbf{40.45} & \textbf{14.17} & \textbf{21.84} & \textbf{0.6027} & \textbf{-2.362} \\
Ablation \cite{urlana2022tesum}        & 39.30          & 13.65          & 21.42          & 0.5967          & -2.457          \\ \Xhline{3\arrayrulewidth}
\end{tabular}
}
\caption{Performance comparison of the Multi-News and \textsc{Multi-News\textsuperscript{+}} datasets on two models. The “Ablation” row represents a version of the Multi-News dataset that has been cleansed using methods from previous study \cite{urlana2022tesum}.}
\label{tab-main}
\end{table*}

The previous Multi-News dataset plays an important role in multi-document summarization research. It consists of sets of documents and their corresponding summaries. However, as shown in Table~\ref{tab-example} and detailed in Appendix~\ref{sec:appendix-example} and \ref{sec:appendix-extreme}, the Multi-News dataset contains several noisy and irrelevant articles that are unrelated to the summary or other documents. This issue arises from their construction process, which relies on automated crawling from the Internet Archive.

To solve this issue and cleanse the dataset, we defined our problem as a classification task determining whether each document is relevant to the summary. To this end, we designed the prompt for the model as shown in Appendix~\ref{sec:appendix-prompt}. We integrated CoT to enhance the model's performance by evaluating the relevance of each document to the summary. Thus, a \textit{rationale} for the decision can be made available, which marks the difference between LLM-based and human annotations. While traditional human annotation through crowdsourcing platforms like Amazon Mechanical Turk usually produces annotation results without underlying reasons due to additional costs, LLM-based annotators can easily offer explanations through CoT. These rationales can assist human managers in reviewing results and rectifying erroneous decisions.

Furthermore, we imitated the conventional dataset cleansing procedure which typically involves multiple human annotators and their collective judgments, primarily through majority voting. Similarly to the majority voting process used by human annotators, we applied this approach to the LLM-based annotators. In particular, we generated five individual LLM agents to read the summary and documents and determine if the document is relevant to the summary. This strategy based on self-consistency can boost the quality of annotations, by rectifying potential errors made by individual agents \cite{wang2022self}. Figure~\ref{fig-framework} presents the summary of the overall process.

Based on the proposed method, we utilized five LLM agents to individually annotate 56,216 sets of summaries and documents from the Multi-News dataset. Specifically, we employed the \texttt{GPT-3.5-turbo-0125} model\footnote{\texttt{GPT-3.5-turbo-0125} charges 0.0005\$ for the input of 1,000 tokens, and 0.0015\$ for the generation of 1,000 tokens.}, the most recent model at the time of this study. With a prompt designed for a 3-shot CoT, approximately 3,500 tokens were required to annotate the input summaries and articles, along with around 100 tokens for generating reasoning processes and annotation results. The cost per annotation sample amounted to approximately 0.01\$ (0.002\$ per agent), resulting in a total cost of approximately 550\$ to annotate the entire Multi-News dataset.

After annotation, we found that 27,052 of the 153,091 articles can be considered noisy documents and do not contribute to the summarization. Subsequently, we constructed \textsc{Multi-News\textsuperscript{+}} by removing these noisy documents from Multi-News while preserving the train/valid/test split. Figure~\ref{fig-histogram} presents the comparison of the Multi-News and \textsc{Multi-News\textsuperscript{+}} datasets in terms of the number of documents per set. More than 15\% of the documents in Multi-News are irrelevant, diminishing the dataset’s quality and degrading the model’s performance. Furthermore, 379 sets have no relevant source articles, as shown in Appendix~\ref{sec:appendix-extreme}. In contrast, by deleting noisy documents, \textsc{Multi-News\textsuperscript{+}} demonstrates enhanced quality.

\section{Experiment}
\subsection{Experimental Design}
To validate the efficacy of data cleansing and the development of \textsc{Multi-News\textsuperscript{+}} in filtering out noisy documents and improving the performance of downstream task models, we measured the multi-document summarization performance of models trained on each dataset, similar to previous study \cite{guo2022questioning}. Enhanced model performance indicates superior dataset quality \cite{ye2022zerogen, choi2024gpts}. We fine-tuned two different models, BART \cite{lewis2020bart} and T5 \cite{raffel2020exploring} on Multi-News and \textsc{Multi-News\textsuperscript{+}}. Performance evaluation metrics included the following metrics: ROUGE \cite{lin2004rouge}, BERTScore \cite{zhang2020bertscore}, and BARTScore \cite{yuan2021bartscore}. For a fair comparison, we used the test set of \textsc{Multi-News\textsuperscript{+}} for each model and reported the average performance across three random seeds.

\subsection{Result}
The results in Table~\ref{tab-main} demonstrate the superiority of the \textsc{Multi-News\textsuperscript{+}} dataset in enhancing the performance of summarization models compared to the original Multi-News dataset. Across various metrics, models trained on \textsc{Multi-News\textsuperscript{+}} consistently outperform those trained on Multi-News, indicating better summarization quality with the refined dataset. This highlights the effectiveness of dataset cleansing in removing noisy and irrelevant documents, thereby enhancing the overall performance of summarization models. Additionally, we performed a human evaluation on the output of 379 sets that are classified as having no relevant source articles and found that 356 sets are correctly classified, which represents 93.9\% of the human-machine agreement rate. We provide an example of error analysis in Appendix~\ref{sec:appendix-error}.

Additionally, we conducted an ablation study using the cleansing method proposed by a previous study \cite{urlana2022tesum}, detailed in Appendix~\ref{sec:appendix-tesum}. Our findings indicate that this method is ineffective in improving downstream task performance on the Multi-News dataset, which focuses on multi-document summarization and differs from the configuration used in the prior study. This underscores the effectiveness of our proposed method and the value of \textsc{Multi-News\textsuperscript{+}}.

\section{Discussion and Future Works}
\label{sec:discussion}
In this section, we discuss recent advancements in the field since the submission of the manuscript and propose strategies for incorporating them in future research.

\noindent \textbf{Cutting-edge models.} Although we employed five \texttt{GPT-3.5-turbo-0125} models for our experiments, the field has seen the release of more advanced models, such as \texttt{GPT-4o} \cite{openai2024gpt4o}, \texttt{GPT-4o-mini} \cite{openai2024gpt4omini}, and \texttt{OpenAI O1} \cite{openai2024o1}, along with the continued development of open-source models like \texttt{LLaMA-3} \cite{dubey2024llama}, \texttt{Gemma-2} \cite{team2024gemma}, and \texttt{Mistral Nemo} \cite{mistral2024nemo}. Models such as \texttt{GPT-4o-mini} and other open-source alternatives offer reduced costs compared to \texttt{GPT-3.5-turbo-0125}, making their adoption promising for both lowering the expense of dataset cleansing and improving the accuracy of detecting noisy documents.

\noindent \textbf{Weighted majority voting.} The availability of high-performance yet cost-effective models like \texttt{GPT-4o} presents the opportunity to use them as \textit{expert} annotators, given their superior capabilities compared to models like \texttt{GPT-3.5-turbo-0125} or \texttt{GPT-4o-mini}. For example, rather than using five \texttt{GPT-3.5-turbo-0125} models, we could employ three \texttt{GPT-3.5-turbo-0125} models alongside one \texttt{GPT-4o}, with \texttt{GPT-4o} carrying double the weight of a \texttt{GPT-3.5-turbo-0125} annotator. This approach positions \texttt{GPT-4o} as an expert, where agreement between at least one \texttt{GPT-3.5-turbo-0125} model and \texttt{GPT-4o} would trigger document deletion.

\noindent \textbf{Supervision from superior models.} Another potential approach involves using more capable models to verify annotation results. In this scenario, \texttt{GPT-4o} would not participate in the initial annotation process but would instead verify the outcomes produced by \texttt{GPT-3.5-turbo-0125} models. By taking the documents, summaries, and annotation results as input, \texttt{GPT-4o} acts as an expert reviewer overseeing the outputs of standard annotators.

\noindent \textbf{Cost-efficient cleansing via pre-screening.} In this paper, we applied the data cleansing strategy to every document in the dataset. However, a more cost-efficient approach could involve performing the annotation procedure only on documents likely to contain noise. Techniques such as dataset cartography \cite{swayamdipta2020dataset} could serve as a pre-screening method to identify cleansing candidates, thereby reducing the overall cost of dataset cleansing.

\section{Conclusion} 
In this study, we suggest deploying cost-efficient LLM-based data annotation to cleanse real-world datasets by identifying and excluding irrelevant and noisy data. We conducted a case study using this strategy to cleanse the Multi-News dataset and proposed the improved \textsc{Multi-News\textsuperscript{+}} dataset. Our case study revealed that \textsc{Multi-News\textsuperscript{+}} provides superior data quality compared to the original Multi-News dataset. Additionally, we have made \textsc{Multi-News\textsuperscript{+}} publicly available, thereby supporting further research in the field of multi-document summarization.

Our work paves the road to extending our data cleansing strategy to other datasets, broadening the scope of utilizing LLMs. This extension would enhance the quality of existing datasets across various domains without the need to construct new datasets from scratch. As such, our approach not only contributes to the advancement of multi-document summarization research but also offers a cost-efficient solution for enhancing dataset quality. We are committed to extending our LLM-based method to other datasets, further solidifying its applicability to other tasks.

\section*{Limitations}
We acknowledge several limitations regarding our proposed method. First, our method is primarily limited by the possibility of wrong classification even with majority voting and CoT. In the future, we may adopt various LLMs as agents and apply weighted majority voting according to their performance to alleviate this issue, as discused in Section~\ref{sec:discussion}.

Secondly, the nature of the Multi-News dataset might exhibit a real-world case of automatic collection of documents from the web that are not always relevant to the summary. In other words, the inclusion of noisy documents might demonstrate the characteristics of real-world automatic crawling. For instance, the model trained on the Multi-News dataset may be more suitable for a real-time system that automatically crawls data from the web and summarizes them. However, we believe such a possibility can be dealt with through the reciprocal usage of our \textsc{Multi-News\textsuperscript{+}} and previous Multi-News dataset. For instance, one could utilize a previous Multi-News dataset when the trained model is expected to consistently deal with noisy documents for inference and there are no pre-defined strategies for filtering out these noisy documents at inference time. Otherwise, for cases where the model is expected to only handle clean documents, it will be more beneficial to utilize our proposed \textsc{Multi-News\textsuperscript{+}} dataset for training the model.

\section*{Ethics Statement}
As we are exploiting LLMs for classifying irrelevant documents rather than text generation, the ethical concern with our method is smaller than that of studies that utilize LLMs to generate texts. Nonetheless, recent studies suggest that the CoT technique may induce ethical bias in LLM \cite{shaikh2023second}. In future work, we plan to investigate this phenomenon's appearance in our method.

\section*{Acknowledgements}
This research was supported by Basic Science Research Program through the National Research Foundation of Korea(NRF) funded by the Ministry of Education(NRF-2022R1C1C1008534), and Institute for Information \& communications Technology Planning \& Evaluation (IITP) through the Korea government (MSIT) under Grant No. 2021-0-01341 (Artificial Intelligence Graduate School Program, Chung-Ang University).

\bibliography{custom}
\bibliographystyle{acl_natbib}

\newpage
\appendix

\section{Dataset Statistics}
\label{sec:appendix-stats}
\textsc{Multi-News\textsuperscript{+}} keeps the train/valid/test split of Multi-News, which is 80\%, 10\%, and 10\%. Table~\ref{tab-stats} displays the number of articles per each split.

\begin{table}[h]
\centering
\resizebox{\columnwidth}{!}{%
\begin{tabular}{c|cc|cc|cc}
\Xhline{3\arrayrulewidth}
           & \multicolumn{2}{c|}{Multi-News}     & \multicolumn{2}{c|}{\textsc{Multi-News\textsuperscript{+}}} & \multicolumn{2}{c}{\% of modification} \\
           & Sets           & Articles           & Sets                       & Articles                       & Sets             & Articles            \\ \hline\hline
Train      & 44,972         & 125,417            & 44,668                     & 102,057                        & 0.7\%            & 18.6\%              \\
Validation & 5,622          & 15,367             & 5,585                      & 12,509                         & 0.7\%            & 18.6\%              \\
Test       & 5,622          & 15,505             & 5,584                      & 12,703                         & 0.7\%            & 18.1\%              \\ \Xhline{3\arrayrulewidth}
\end{tabular}
}
\caption{Number of sets and articles per each split.}
\label{tab-stats}
\end{table}

\section{Construction Process of Multi-News}

\begin{figure}[t!]
    \centering
    \includegraphics[width=\columnwidth]{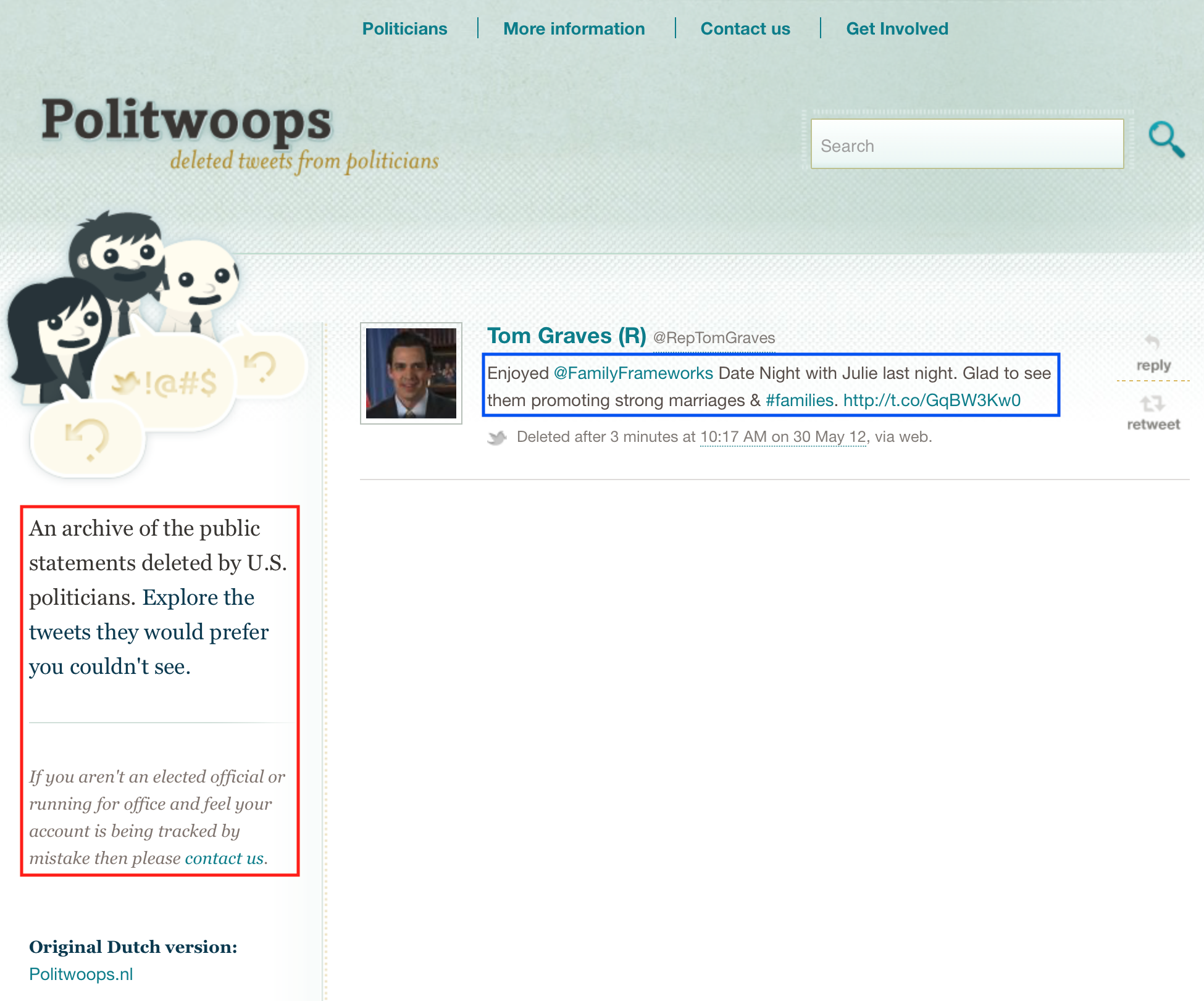}
    \caption{A screenshot of a webpage that is relevant to the article in Appendix~\ref{sec:appendix-extreme}. Multi-News includes the text in the red box instead of the desired content in the blue box.}
\label{fig-screenshot}
\end{figure}


In this section, we briefly explain the construction process of the Multi-News dataset. Multi-News is based on data from newser.com\footnote{\url{https://newser.com}} that offers human-written summaries of news articles. Each summary is written by professional human editors and involves several outlinks to the original articles and relevant websites. Multi-News collected this human-written summary and documents from its outlinks, which behave as source documents for summarization. Notably, the authors of Multi-News archived every article leveraging Wayback Machine\footnote{\url{https://web.archive.org}}, a system that supports archiving of the circumstances of a given website, to ensure the reproducibility and support future investigation. Contents of each document have been accessed and crawled from these Wayback-archived links. 

However, this affected problems regarding the quality of the dataset. As shown in examples of noisy documents in Appendix~\ref{sec:appendix-example}, several noisy documents consist of a message from Wayback Machine. Moreover, the failure to crawl the content of the webpage caused other problems. We investigated the case shown in Appendix~\ref{sec:appendix-extreme} and found that it is a result of the crawling of the wrong part of the website. Figure~\ref{fig-screenshot} clearly showcases this phenomenon where the content in the red box is crawled instead of the content in the blue box, which is desired. Even though the content in the blue box is different for each article, the system wrongly crawled the shared red box, which resulted in five noisy documents that share the same content and do not contribute to the summary.

From the example above, we revealed the presence of the wrongly crawled documents, that affect the quality of the dataset. We believe such phenomena would be alleviated with the advancement of LLM-based autonomous agents \cite{wang2023survey}, as they could visit the website and only crawl the text relevant to the summary. Even though we leave this as future work, this research direction should be prompted.

\section{Implementation Details}
We utilized PyTorch \cite{paszke2019pytorch} and Huggingface Transformers \cite{wolf2020transformers} to implement and evaluate the model. Specifically, we employed \textit{facebook/bart-large-cnn}\footnote{Note that this model is already fine-tuned with the CNN/DM dataset \cite{nallapati2016abstractive}, a single-document summarization dataset.} and \textit{google-t5/t5-base}, with 406M and 220M parameters, respectively, for BART and T5. Each model was trained using Adam \cite{kingma2015adam} with a learning rate of 2e-5 over 3 epochs. We used a batch size of 4 and implemented a gradient accumulation step of 4, resulting in a practical batch size of 16. For evaluation, we utilized \textit{bert-base-uncased} and \textit{facebook/bart-large-cnn} for BERTScore and BARTScore, respectively. We reported BERTScore-F1 in Table~\ref{tab-main}. ROUGE scores were measured using the rouge-score\footnote{\url{https://pypi.org/project/rouge-score/}} library, with the F1 score of each metric. The training was conducted on a single NVIDIA A100 40GB GPU. We provide the source code and dataset to the public.\footnote{\url{https://github.com/c-juhwan/multi_news_plus}}

For the human evaluation, we recruited three volunteers and individually asked them to determine whether the decision of the model was correct or not given the summary, original articles, and rationale of the model. We defined the model made an incorrect decision when at least one human evaluator flagged the output as an incorrect classification.

\begin{table}[t!]
\centering
\resizebox{\columnwidth}{!}{%
\begin{tabular}{c|cc}
\Xhline{3\arrayrulewidth}
Model                                  & \multicolumn{2}{c}{\textit{Mistral-7B-Instruct-v0.2}}                                          \\
Metric                                 & BERTScore       & BARTScore       \\ \hline\hline
No Noisy Example                       & 0.6004          & -2.704          \\
One Noisy Example                      & 0.5976          & -2.721          \\
Two Noisy Examples                     & 0.5954          & -2.738          \\ \Xhline{2\arrayrulewidth}
Model                                  & \multicolumn{2}{c}{\textit{Llama-2-7b-chat-hf}}                                \\
Metric                                 & BERTScore       & BARTScore       \\ \hline\hline
No Noisy Example                       & 0.6038          & -2.507          \\
One Noisy Example                      & 0.6022          & -2.521          \\
Two Noisy Examples                     & 0.6016          & -2.539          \\ \Xhline{3\arrayrulewidth}
\end{tabular}
}
\caption{Performance of LLM-based summarization of Multi-News with different amounts of noisy examples. We only report two model-based metrics as the human-generated reference summary has a different form compared to the LLM-generated summary.}
\label{tab-llm}
\end{table}

\section{Manual Analysis}
\label{sec:appendix-manual}

To perform a more detailed analysis of the accuracy of the proposed method, we randomly selected 60 instances from the validation set, which comprises 153 documents. A confusion matrix was defined to evaluate the classification for each document as follows:

\begin{itemize}
    \item True Positive (TP): Relevant documents that were correctly classified as relevant.
    \item False Positive (FP): Documents classified as relevant but are not actually relevant.
    \item True Negative (TN): Irrelevant documents correctly classified as not relevant.
    \item False Negative (FN): Relevant documents incorrectly classified as not relevant.
\end{itemize}

Upon review, we found that 127 documents were classified as TP, 24 as TN, and 2 as FN. The annotation framework identified 26 documents as irrelevant and noisy, which accounts for approximately 17\% of the total 153 documents. This aligns closely with the statistics in Table~\ref{tab-stats} of Appendix~\ref{sec:appendix-stats}, which indicates that 18.6\% of documents in the validation set were classified as noisy.

From these results, the precision is 1.0, as there were no FP documents, while the recall is approximately 0.984. Additionally, we observed that 17 of the 24 TN documents could be classified as noisy system messages, such as ``This will appear next to all of your comments; this will not appear anywhere on Newser,'' as illustrated in Appendix~\ref{sec:appendix-example}. The remaining 7 documents were irrelevant to the summary.

Furthermore, we investigated the two FN cases. In one instance, the summary included a portion related to the misclassified document at the very end. In the other, the misclassified document provided context for the summary but was not directly connected to it. These cases are consistent with the error patterns discussed in Appendix~\ref{sec:appendix-error}.

It is important to note that while individual annotators occasionally made incorrect classifications, the majority voting process effectively corrected these errors. This highlights the efficacy of our proposed method in improving data annotation quality and ensuring thorough dataset cleansing.

\section{Additional Experiment with Large Language Models}
\label{sec:appendix-llm}

This section introduces our additional experiment that investigates the influence of noisy examples for LLMs in a few-shot learning scheme. For this purpose, we used 7B-sized, instruction-tuned Llama2 \cite{touvron2023llama} and Mistral \cite{jiang2023mistral}. Specifically, we used \textit{meta-llama/Llama-2-7b-chat-hf} and \textit{mistralai/Mistral-7B-Instruct-v0.2} from Transformers \cite{wolf2020transformers}. In this experiment, we prompted the model to summarize the documents in the test set of Multi-News with two-shot examples selected from the training set of Multi-News. Additionally, we differentiated the number of noisy documents in the examples given as the prompt. Table~\ref{tab-llm} presents the experimental result. The result demonstrates that the inclusion of the noise in the example degrades the quality of the summary generated by the LLM. This suggests the significance of the exclusion and filtering of the noise for LLMs, which underscores the necessity of dataset cleansing presented in this paper.

\section{Analysis of Multi-News}
\label{sec:appendix-tesum}

\begin{table}[h]
\centering
\resizebox{0.8\columnwidth}{!}{%
\begin{tabular}{c|c}
\Xhline{3\arrayrulewidth}
                           & Multi-News  \\ \hline
Dataset Size               & 56,216      \\ \hline
Source Article Size        & 156,289     \\ \hline
Avg Words in Source        & 433.62      \\ \hline
Avg Sentences in Source    & 23.42       \\ \hline
Avg Words in Summary       & 228.69      \\ \hline
Avg Sentences in Summary   & 11.52       \\ \hline\hline
Empty Summary              & 0           \\ \hline
Duplicated Summary         & 0           \\ \hline
Prefixes Summary           & 0           \\ \hline
Empty Source               & 570         \\ \hline
Duplicated Source          & 544         \\ \hline
Source < 4 Sentences       & 45          \\ \hline
Source < 40 Words          & 7           \\ \hline
Summary < 10 Words         & 0          \\ \hline
Compression < 50\%         & 31,994      \\ \hline
Compression > 80\%         & 390         \\ \hline
Abstractivity < 10         & 496         \\ \hline
Abstractivity > 80         & 126         \\ \hline\hline
Avg Abstractivity          & 41.42       \\ \hline
Avg Compression            & 46.19\%     \\
\Xhline{3\arrayrulewidth}
\end{tabular}
}
\caption{The result of analysis of Multi-News dataset with rule-based filtering methods \cite{urlana2022tesum}. We concatenated every source document to measure their average word and sentence length.}
\label{tab-analysis}
\end{table}

Following the previous study of TeSum \cite{urlana2022tesum}, we apply filtering strategies and analyze the characteristics of Multi-News with these strategies. Table~\ref{tab-analysis} exhibits the result of the analysis. First, we found that 0.7\% of total source documents can be considered noisy documents as it is empty or duplicated from other source documents within the same set. Second, we found previous rule-based filtering methods are not very effective standards for the Multi-News dataset. For instance, there were no sets that had empty summaries, summaries that were duplicated with other summaries, or summaries that repeated the first few sentences of source documents. The only exception is \textit{Compression < 50\%}, which identified more than half of the dataset. However, it should be noted that Multi-News is a multi-document summarization dataset, which is different from datasets for previous studies. For instance, average compression is significantly lower than other single-document summarization datasets reported in the previous study \cite{urlana2022tesum}, as multiple source documents in Multi-News involve more information compared to the source document of single-document summarization datasets. In conclusion, this analysis demonstrates that previous filtering strategies are less practical for multi-document summarization datasets such as Multi-News and enlightens the necessity of novel approaches for these datasets.

\newpage
\onecolumn
\section{Examples of Noisy Documents}
\label{sec:appendix-example}
This section demonstrates several examples of noisy documents observed in the Multi-News dataset not related to the summary. Please refer to the released dataset file for details. \newline

\setlength{\parindent}{0cm}
\begin{minipage}[t]
{\linewidth}\raggedright
\xhrulefill[thickness=1pt]
\vspace{1mm}

\begin{itemize}
\item Tweet with a location you can add location information to your tweets, such as your city or precise location, from the web and via third-party applications. You always have the option to delete your tweet location history. Learn more
\item Focused crawls are collections of frequently-updated webcrawl data from narrow ( as opposed to broad or wide ) web crawls, often focused on a single domain or subdomain.
\item Dow jones reprints: this copy is for your personal, non-commercial use only. To order presentation-ready copies for distribution to your colleagues, clients or customers, use the order reprints tool at the bottom of any article or visit www.djreprints.com
\item This crawl of online resources of the 115th us congress was performed on behalf of the united states national archives \&amp; records
\item The seed for this crawl was a list of every host in the wayback machine this crawl was run at a level 1 ( urls including their embeds, plus the urls of all outbound links including their embeds ) the warc files associated with this crawl are not currently available to the general public.
\item These crawls are part of an effort to archive pages as they are created and archive the pages that they refer to. That way, as the pages that are referenced are changed or taken from the web, a link to the version that was live when the page was written will be preserved.then the internet archive hopes that references to these archived pages will be put in place of a link that would be otherwise be broken, or
\item Please enable cookies on your web browser in order to continue. The new european data protection law requires us to inform you of the following before you use our website: we use cookies and other technologies to customize your experience, perform analytics and deliver personalized advertising on our sites, apps and newsletters and across the internet based on your interests. By clicking ``i agree'' below, you consent to the use by us and our third-party partners of cookies and data gathered from your use of our platforms. See our privacy policy and third party partners to learn more about the use of data and your rights. You also agree to our terms of service.
\item Thank you for reading. Please purchase a subscription to continue reading. A subscription is required to continue reading. Thank you for reading 5 free articles. You can come back at the end of your 30-day period for another 5 free articles, or you can purchase a subscription and continue to enjoy valuable local news and information. If you are a current 7-day subscriber you are granted an all-access pass to the website and digital newspaper replica. Please click sign up to subscribe, or login if you are already a member. Thank you for reading 5 free articles. You can come back at the end of your 30-day period for another 5 free articles, or you can purchase a subscription and continue to enjoy valuable local news and information. If you are a current 7-day subscriber you are granted an all-access pass to the website and digital newspaper replica. Please click below to get started.
\item Add a location to your tweets when you tweet with a location, twitter stores that location. You can switch location on/off before each tweet and always have the option to delete your location history. Learn more
\end{itemize}

\vspace{1mm}
\xhrulefill[thickness=1pt]
\end{minipage}

\newpage
\section{Extreme Cases of Noisy Documents}
\label{sec:appendix-extreme}
In addition to examples of noisy documents, we discovered the following extreme case of noisy data in the Multi-News dataset. In this example, five documents have the same content but offer no information on the summary. Thus, it cannot generate a reasonable summary based on the given documents. We witnessed 379 similar cases during the dataset cleansing process, as reported in Figure~\ref{fig-histogram}. While they were excluded from training and testing, we included them in the dataset file for future investigation. \newline

\begin{minipage}[t]
{\linewidth}\raggedright
\xhrulefill[thickness=1pt]
\vspace{1mm}

\texttt{Summary} \newline 
Note to tweeting politicians: watch what you post, because politwoops will remember it forever. The transparency-minded website is safeguarding politicians'deleted tweets, enabling the rest of us to giggle or ponder over them at our leisure, the atlantic reports. The site's current 6-month stash includes a few doozey deletions, including john mccain mocking vladimir putin's tears and rep. Jeff miller posting a link to a poll that asked, " was obama born in the united states? " a few deletions are more odd than obvious, begging us to ask what politicians were thinking. Why, for example, did rep. Tom graves remove a tweet about going out one night with his wife? or rep. Kathy hochul delete one about her visit to a cancer institute? perhaps rep. Stephen fincher's tweet comparing the bachelor to the hunger games is a more obvious case, but the online avenues of a politician's mind can be dimly lit indeed. \newline\newline

\texttt{Document 1} \newline 
An archive of the public statements deleted by u.s. Politicians. Explore the tweets they would prefer you couldn't see. If you aren't an elected official or running for office and feel your account is being tracked by mistake then please contact us. \newline\newline

\texttt{Document 2} \newline 
An archive of the public statements deleted by u.s. Politicians. Explore the tweets they would prefer you couldn't see. If you aren't an elected official or running for office and feel your account is being tracked by mistake then please contact us. \newline\newline

\texttt{Document 3} \newline 
An archive of the public statements deleted by u.s. Politicians. Explore the tweets they would prefer you couldn't see. If you aren't an elected official or running for office and feel your account is being tracked by mistake then please contact us. \newline\newline

\texttt{Document 4} \newline 
An archive of the public statements deleted by u.s. Politicians. Explore the tweets they would prefer you couldn't see. If you aren't an elected official or running for office and feel your account is being tracked by mistake then please contact us. \newline\newline

\texttt{Document 5} \newline 
An archive of the public statements deleted by u.s. Politicians. Explore the tweets they would prefer you couldn't see. If you aren't an elected official or running for office and feel your account is being tracked by mistake then please contact us. 

\vspace{1mm}
\xhrulefill[thickness=1pt]
\end{minipage}

\newpage
\section{Error Analysis}
\label{sec:appendix-error}
Following the form of the previous study \cite{choi2024gpts}, we provide an error analysis to provide a more balanced view of the behavior and limitations of our proposed method. In the first example, we can observe that while Document 1 can be regarded as irrelevant to the summary except that there is a mention of fusion tv, Document 2 contains information about Mike Tyson and his new TV documentary series. However, the model predicted both documents are irrelevant to the summary, primarily because the model concentrated on the mention of the ``world team tennis exhibition'' from Document 2. From this insight, we hypothesize GPT-3.5 suffers from a mixture of irrelevant and relevant information in one document.

\begin{minipage}[t]
{\linewidth}\raggedright
\xhrulefill[thickness=1pt]
\vspace{1mm}

\texttt{Summary} \newline
Over his career, former heavyweight champion mike tyson recorded 50 wins and six losses. But he recently notched another big loss in latin america — this time as a coach of a bird, reports the ap. Tyson traveled to suriname as part of the new fusion tv documentary series outpost, and was soundly beaten when he entered a bird in a songbird contest, a cherished local tradition. Cameras captured iron mike as he learned about the contest, located a bird to enter — he dubbed the tiny guy " little mike " — but then suffered a tko when a competing champion cheeped and peeped more than his bird did in the same 15-minute period. " little mike let us down, man. I was in his corner, though, " said tyson. " it was just amazing meeting the people, meeting the culture — i had a great time. " the series, kicking off on sunday with tyson's episode, mixes travel adventure, history, and journalism to shine a light on global stories. The first season focuses on latin america and includes as hosts the late show with stephen colbert bandleader jon batiste, brain games star jason silva, and transgender model carmen carrera. Spanish versions air on unimas. Tyson was lured onto the show by the chance to visit a country he'd never heard of and his love of birds. The former boxer has loved pigeons and kept them since he was a kid in brooklyn. ( sunday's show recorded the moment tyson lovingly released his bird in suriname. ) " my wife always says the reason i keep my pigeons is they connect me to my childhood, " tyson said. " once it's in your blood, it never leaves. It's just who you are. " \newline

\texttt{Document 1} \newline
Starting in 1996, alexa internet has been donating their crawl data to the internet archive. Flowing in every day, these data are added to the wayback machine after an embargo period. \textit{[Abbreviated duplicated text]} Outpost shows you the world like you've never seen it. The series lives at the intersection of investigative journalism and adventure travel, bringing you a local perspective on faraway places and inviting you to explore. The series premieres march 26 @ 8 and 11 pm on fusion tv. In the first episode, transgender model carmen carrera travels to brazil, a place where rates of violence against lgbt people are some of the highest in the world, to find out what's happening, what life is like for young transgendered people in brazil, and what the future might hold. Gabriel leigh takes us to el alto, bolivia, where some of the craziest architecture on earth is taking shape as part of a surge in indigenous purchasing power. \newline

\texttt{Document 2} \newline
\textit{[Abbreviated duplicated text]} file - in this monday, oct. 10, 2016, file photo, mike tyson attends a world team tennis exhibition to benefit the elton john aids foundation in las vegas. Tyson traveled to suriname as part of the new fusion tv documentary series "outpost " and was soundly beaten when he entered a bird in a songbird... ( associated press ) \textit{[Abbreviated duplicated text]} new york ( ap ) — over his career, former heavyweight champion mike tyson recorded 50 wins and six losses. But he recently notched another big loss in latin america — this time as a coach of a bird. Tyson traveled to suriname as part of the new fusion tv documentary series " outpost " and was soundly beaten when he

\vspace{1mm}
\xhrulefill[thickness=1pt]
\end{minipage}

\newpage

This second example also showcases the characteristics of GPT-3.5 model we used. In this example, it is obvious that Document 2 is less relevant to the summary, which is mainly about the relationship between Gwyneth Paltrow and Chris Martin. However, while it is not the main content of the document as well as Document 2, Document 1 contains a sentence that mentions the relationship between the two (``her amicable split from husband chris martin of coldplay''). Nonetheless, the model predicted Document 1 is also irrelevant to the summary, implying the model is stringent to the \textit{partial} contribution of the document to the summary. However, it is important to note that we categorized these instances as errors based on rigorous human evaluation, and such errors constituted fewer than 10\% of the total classifications, where a single flag by multiple human evaluators was sufficient to deem it an error. We are planning to manually revise these errors in the released version of \textsc{Multi-News\textsuperscript{+}}.

\begin{minipage}[t]
{\linewidth}\raggedright
\xhrulefill[thickness=1pt]
\vspace{1mm}

\texttt{Summary} \newline
Gwyneth paltrow continues to paint the sunniest of pictures of her post-conscious-uncoupling life with chris martin, but the description she gives glamour in a new interview may be the most interesting one so far. " we're still very much a family, even though we don't have a romantic relationship. He's like my brother, " she says, explaining that the two of them and their two kids still spend quite a bit of time together, even staying in one another's houses and spending holidays together ( not to mention collaborating on songs together ). " the ideal is to stay married. But if you can't stay married, wouldn't the ideal be that you could still be a family and you could put aside your own stuff long enough to explore — what is this new family and who am i in it? " paltrow muses. " and chris is a great ex-husband ' cause he's a very, very willing partner in how to do that. " she adds that, though she's " very independent, " she does see the value in having a husband, and though she's not quite divorced yet, she could perhaps see herself getting married again someday. ( click to see what she has to say about her other famous exes. ) \newline

\texttt{Document 1} \newline
Gwyneth paltrow is in a state of deep focus. The new goop office is under construction — "it's like a dust bowl, " she says with a laugh — so today she's helming her company from the kitchen island of her los angeles home. Fitting, considering it was at her kitchen table ( then in london ) that paltrow, 43, started goop as a newsletter to friends nearly eight years ago. Since then, she has built goop into a global brand: it has produced sought-after collaborations with valentino and stella mccartney; opened pop-up shops; and brought terms like conscious uncoupling and vaginal steaming to the masses ( the first a description of her amicable split from husband chris martin of coldplay; the second, a way to cleanse one's uterus — don't try it at home ). Her presence has also unwittingly exposed a dirty little secret: as fans, we provide actresses with wealth and fame, only to scoff when they actually lead that rich and famous lifestyle publicly. We want these stars to be "just like us. " but paltrow's life simply isn't. She won't pretend that she shops at the dollar store for beauty products or feeds her kids, apple, 11, and moses, 9, a steady diet of fast food; \newline

\texttt{Document 2} \newline
Gwyneth paltrow was definitely in the mood to share during her appearance on howard stern's siriusxm radio show on wednesday.... Especially when it came to her a-list exes. In the hour-long chat, stern of course wanted to know all about paltrow's ex-fiance brad pitt, who the shakespeare in love star was engaged to when she was 24 years old. The beautiful blondes eventually called it quits in 1997 after three years together. Getty images " i didn't think about it at the time, but i ' m sure it did help with my career, " the now 42-year-old actress admits about the start of all the paparazzi attention when the two got together on the set of seven. " i definitely fell in love with him. He was so gorgeous, and sweet -- i mean, he was brad pitt, you know? " video: a history of gwyneth's former flames her parents, the late bruce paltrow and actress blythe danner, also 100 percent approved. " my father was so devastated when we broke up, " she dishes. " my father loved him like he was his son. " in hindsight, she blames the demise of their relationship on her youth. " i was such a kid, i was 22 when we met, " she explains.

\vspace{1mm}
\xhrulefill[thickness=1pt]
\end{minipage}

\newpage
\section{Prompt}
\label{sec:appendix-prompt}
This section describes the prompt used for dataset cleansing. We truncated several parts to save space and only displayed a 1-shot CoT example. Please refer to the released source code for the exact prompt. \newline

\setlength{\parindent}{0cm}
\begin{minipage}[t]
{\linewidth}\raggedright
\xhrulefill[thickness=1pt]
\vspace{1mm}
\newline
\texttt{System} \newline
You are a helpful assistant. Your job is to identify the document that is not relevant to the summary. The relevance is determined by the mention of the key points in the summary. For example, the alert message from the twitter, the location information, and the unrelated information are considered irrelevant. Please response with the resaoning process and the result as document number or `None'. Do not end the response with a period. If multiple documents are irrelevant to the summary, please resonse in the form of `Document 1|Document 3' \newline\newline

\texttt{User} \newline
Given the summary and documents, please identify the document that is not relevant to the summary such as advertisements. If every document is relevant to the summary, output `None'.
Input document will be given in next dialogue. \newline\newline

\texttt{User} \newline
[Summary] \newline
\# The content of summary \newline
[Document 1] \newline
\# The content of document 1 \newline
[Document 2] \newline
\# The content of document 2 \newline
[Document 3] \newline
\# The content of document 3 \newline
[Document 4] \newline
\# The content of document 4 \newline\newline

\texttt{Assistant} \newline
The summary discusses Tucker Carlson's controversial tweet regarding Sarah Palin and the subsequent backlash he faced from conservative women. Document 1 is related to Tucker Carlson's controversial tweet. Document 2 is unrelated as it contains information about adding location information to tweets and does not contribute to the discussion about Tucker Carlson's tweet and the reaction it received. Document 3 is a tweet message that calls Tucker Carlson a misogynist, related to the summary. Document 4 is related to the summary as it deals with Tucker Carlson's movement. Therefore, the irrelevant document is: Document 2

\vspace{1mm}
\xhrulefill[thickness=1pt]
\end{minipage}

\end{document}